\documentclass[runningheads]{llncs}
\usepackage{comment}
\usepackage{hyperref}
\usepackage{graphicx}
\begin{document}
\title{FrameRS: A Video Frame Compression Model Composed by Self supervised Video Frame Reconstructor and Key Frame Selector}

\def\CICAISubNumber{}  
\titlerunning{CICAI2022 submission ID \CICAISubNumber} 
\authorrunning{CICAI2022 submission ID \CICAISubNumber} 
\author{Qiqian Fu\inst{1} \and
Guanhong Wang\inst{1} \and
Gaoang Wang\inst{1}}
\institute{Zhejiang University\\
\email{\{qiqian.21, gaoangwang\}@intl.zju.edu.cn, guanhong@zju.edu.cn}}

\maketitle              

\begin{abstract}
In this paper, we present frame reconstruction model: FrameRS. It consists self-supervised video frame reconstructor and key frame selector. The frame reconstructor, FrameMAE, is developed by adapting the principles of the Masked Autoencoder for Images (MAE) for video context. The key frame selector, Frame Selector, is built on CNN architecture. By taking the high-level semantic information from the encoder of FrameMAE as its input, it can predicted the key frames with low computation costs. Integrated with our bespoke Frame Selector, FrameMAE can effectively compress a video clip by retaining approximately 30$\%$ of its pivotal frames. Performance-wise, our model showcases computational efficiency and competitive accuracy, marking a notable improvement over traditional Key Frame Extract algorithms. The implementation is available on \href{https://github.com/QiqianFu/FrameMAE}{Github}.

\end{abstract}
\section{Introduction}
Since the proposal of the transformer~\cite{vaswani2017attention}, it has achieved tremendous success in the field of NLP, which inevitably leads people to think about how to apply it to the field of computer vision (CV). With the continuous development of the transformer in the CV field, there are now various models that utilize the transformer for different downstream tasks, and they have achieved quite impressive results. Among them, the introduction of Vision Transformer (ViT)~\cite{dosovitskiy2020image} undoubtedly represents a significant advancement of the transformer in the CV field, as its structure breaks the limitations of the transformer in NLP.

\begin{figure}[t]
\hspace*{-2.5cm} 
\includegraphics[width=0.8\paperwidth]{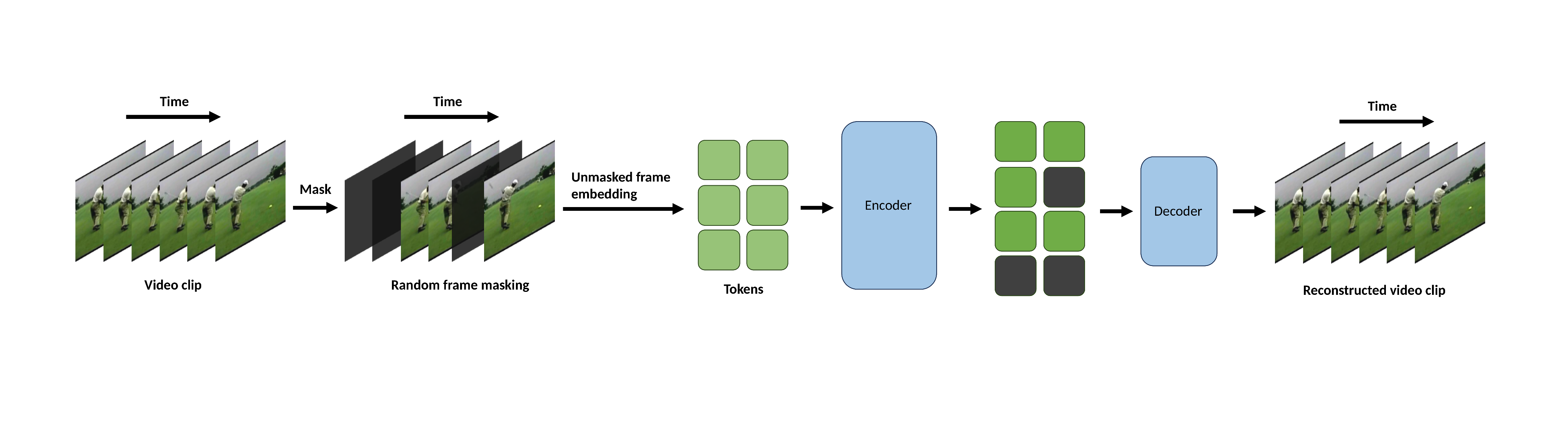}
\caption{The FrameMAE Architecture: We adapted a masking ratio of 37.5$\%$, the unmasked frames will be embedded to tokens, then it is fed into the encoder for processing. The processed tokens will be fed to the decoder together with the masked frames (which is zero matrix) to produce satisfying reconstructed frames.} \label{fig1}
\end{figure}

Masked Autoencoder for Images (MAE)~\cite{he2022masked} adopts the idea of masking tokens in BERT for NLP, transforming an image into multiple tokens and masking some of them, allowing the model to predict the masked tokens. VideoMAE~\cite{tong2022videomae} further extends MAE from individual images to video clips by utilizing tube masking. In our research paper, we proposed a frame masking strategy to train our MAE, which is called FrameMAE (Fig.~\ref{fig1}). This strategy aligns more closely with our objective of reconstructing videos using key frames. For the encoder, we convert the input video into tokens through patch embedding and input them into the encoder. The decoder is then used to reconstruct the pixels. This autoregressive model aligns better with our task.

\begin{figure}
\includegraphics[width=\textwidth]{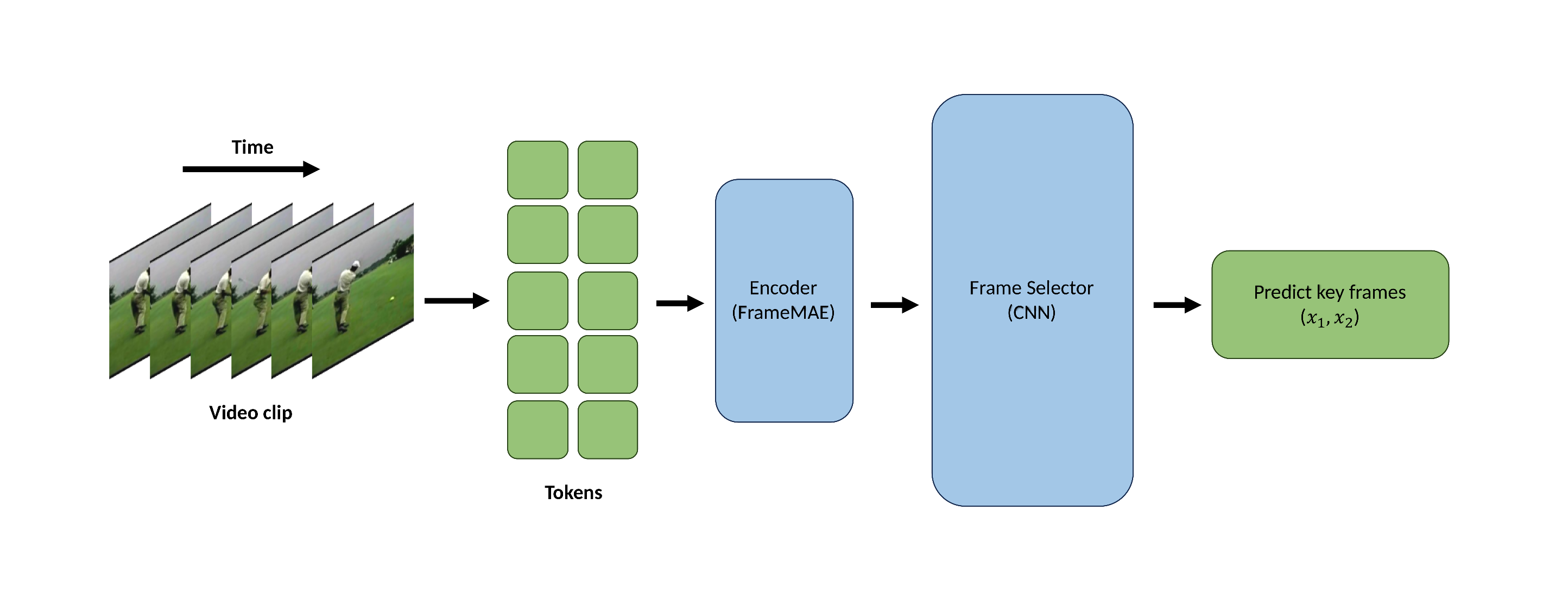}
\caption{The Frame Selector: The design of the Frame Selector make use of the encoder of FrmaeMAE. We leverage the high-level semantic information from the encoder's output as input to the frame selector, thereby reducing computational overhead and enhancing accuracy.} \label{fig2}
\end{figure}

After pretraining an good performing FrameMAE on the SSV2 dataset, we proceeded to design the Frame Selector(Fig.~\ref{fig2}). Since each token in each patch has a dimension of 768, the size of the intermediate layer becomes [768 14 14]. Therefore, we use max pooling to extract the maximum value of each token, reducing the input dimensionality. We then input this into our model, which consists of multiple convolutional layers that we designed. The Frame Selector is trained to predict the 2 frames out of 8 that achieves the best reconstruction results. After training for around 200 epochs, the model can achieve the top-1 accuracy of 27.1\% and the top-5 accuracy of 50.5\%.

\section{Related Work}

\subsection{Vision Transformer} Inspired by Transformer scaling successes in NLP, Dosovitskiy et al.(2020) ~\cite{dosovitskiy2020image} proposed Vision Transformer (ViT), following the original Transformer backbone. The model applies learnable embedding for classification, and positional embedding to retain positional information, which helps ViT to understand image as sentences. It allows Transformer to operate in the field of CV.
\subsection{Masked Language Model} BERT~\cite{devlin2018bert}, which stands for Bidirectional Encoder Representations, designed a customized masked token strategy to pre-train the model. It masks some of the tokens in the original texts randomly and the model is trained to use the context to predict the masked tokens. GPT ~\cite{radford2018improving,radford2019language,brown2020language} is also a Masked Language Model, and its architecture is quite similar to BERT. It uses constrained self-attention while BERT uses bidirectional self-attention. Both designed a successful method for self-supervised per-training.
\subsection{Masked Autoencoder} Image Masked Autoencoders (ImageMAE) , presented by He et al.~\cite{he2022masked}, shows that the concept of masked tokens can also be applied to vision representation learning and it turns out to be effective and high accuracy. Later Feichtenhofer et al. ~\cite{feichtenhofer2022masked} presented an extension model of MAE to spatiotemporal representation learning from videos. It random masked patches of each frames in a video. VideoMAE ~\cite{tong2022videomae} however, adapted tube masking strategy, which means all frames share a same masking map. Both of them have successfully demonstrated that there's a significant amount of redundancy in the video content and Masked Antoencoder is a good way for self-supervised learning.

\section{Proposed Method} 
The FrameRS consists of Video Frame Reconstructor and Key Frame Selector. In this section we will first introduce Video Frame Reconstructor and its pre-training strategy. Then we will introduce Key Frame Selector. Finally, we will demonstrate how to merge them into one video compression model. 
\subsection{Video Frame Reconstructor Pre-training}
The Frame Reconstructor is composed of tranpisformer-based encoder and decoder. Our implementation based on the code of VideoMAE~\cite{tong2022videomae}, which uses tube masking as its pre-training strategy. However, in our project, we replace the tube masking with frame masking. The other settings remain the same. The input video is first divided into non-overlapping patches, then fed into embedding layer to be represented as tokens. The embedding layer is referenced from VideoMAE, and the architecture used is the joint space-time cube embedding described in ViViT ~\cite{arnab2021vivit}. The size of the embedding layer is also configured to be the same as in VideoMAE, which is a stride of [2 16 16] and out channel of 768. Finally, calculate the Mean Square Error between the predicted pixels reconstructed by the decoder with the original pixels and use AdamW to update the parameter.

\subsection{Key Frame Selector}
To address the compression task, a key frame selector is needed. So our motivation is to build a low computation cost model for key frame selecting. The selector is aim to predict the best 4 frames out of 16 with the smallest reconstruction loss. The encoder of FrameMAE is responsible for encoding the source video frames into a a sequence of video features, which is indeed a suitable input for key frame selector. However, the size of the video features is [768, 8*14*14], which is too huge for the selector to deal with. We let it pass through a projector to become 384-dimensional vectors. Since the exact position of the representation is not essential for key frame selecting, it is then fed into a maxpooling layer to decrease the size on the spatial dimension. Then the size is suitable for key frame selector. To reduce the computation cost for key frame selector, we use MLP(Multi-Layer Perception) as the network architecture of key frame selector. The training loss function is cross-entropy loss.

\section{Settings}
\subsection{Data pre-processing}
For FrameMAE pre-training, we randomly sampled 16 frames out of a video clip with a stride of 2 from Something-Something V2. After the video go through the embedding layer, its size will be [768, 8, 196]. Then 3 out of 8 tokens from the same frames will be masked. After completing the pre-training of FrameMAE, we use the pre-trained FrameMAE to generate the dataset for Key Frame Selector. First, we calculate the Mean Square Error of different combination of 2 unmask frames and ranked them from 0 to 27, since there are total 28 possible permutations to choose 2 out of 8 in the temporal dimension. And the ground truth label for each video clip refers to the combination of two that minimizes the loss during reconstruction. And that 2 denote 4 frames from the original video due to the stride of 2.

\section{Experiment}
\begin{figure}[t]
\includegraphics[width=\textwidth]{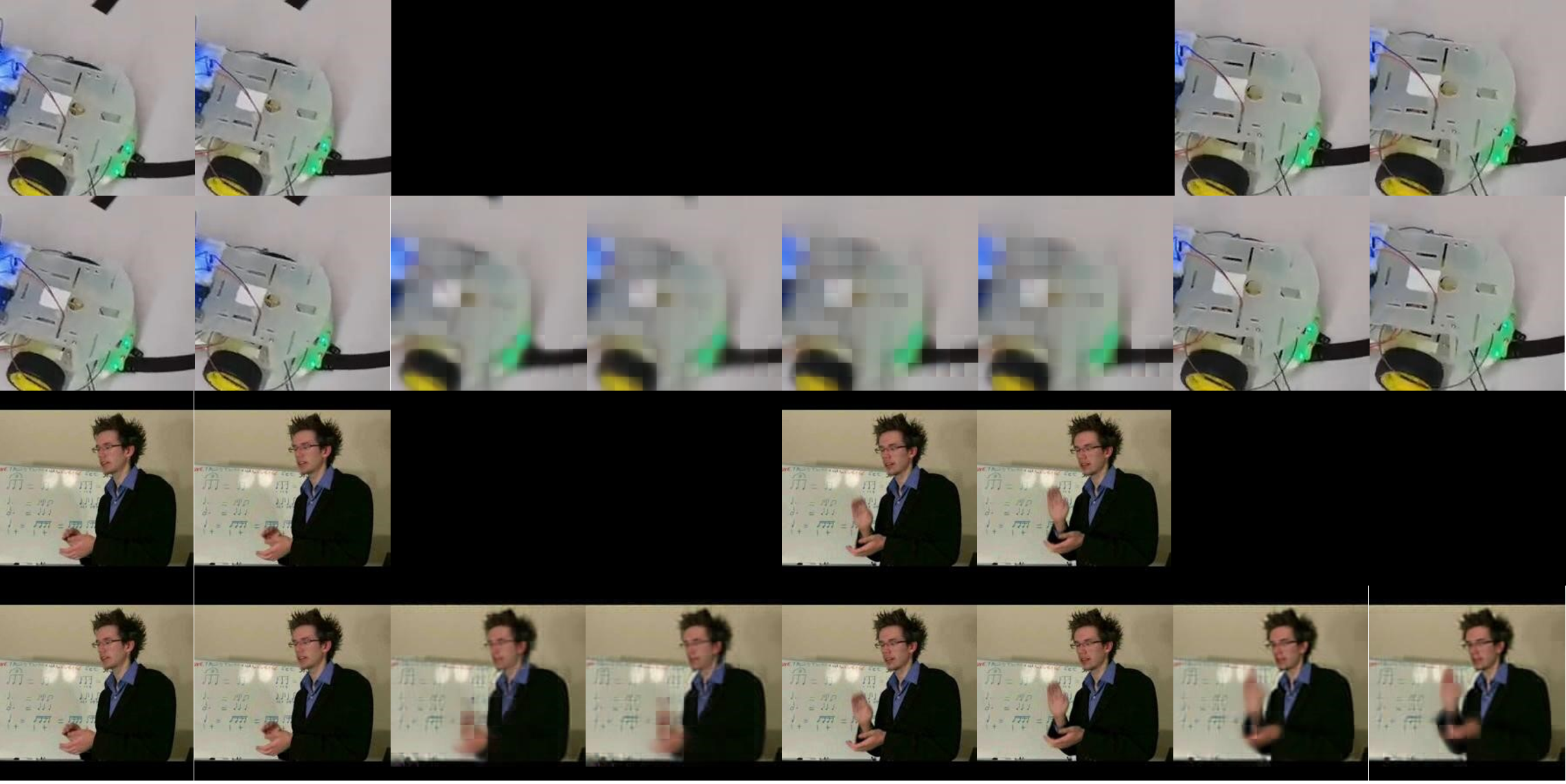}
\caption{The reconstructed results} \label{fig3}
\end{figure}

In this section, we describe the experiments that we use to evaluate the Frame Reconstructor and the Key Frame Selector on different dataset and with different settings. The reconstructed results are shown in Fig.~\ref{fig3}.
\subsection{Datasets}
Most of the experiments are conducted on the Something-Something V2 ~\cite{goyal2017something}. Due to insufficient GPU computational resources, we only use 50k training videos and 10k validation videos for the experiments. 

\subsection{Abolation Study}
We test the  effect of different number of blocks and drop-out rate on accuracy in Table~\ref{tab1}. And it shows that 3 blocks with a drop-out rate of 0.1 perform best. 
\begin{table}[h]
\centering
\caption{The effect of different number of blocks on accuracy.}\label{tab1}
\begin{tabular}{lllll}
blocks & top-1  & top-5   & drop-out & best epoch \\ \hline
3      & 27.1\% & 50.52\% & 0.1      & 224        \\
3      & 26.8\% & 49.7\%  & -        & 204        \\
4      & 25.1\% & 48.6\%  & -        & 200          
\end{tabular}
\end{table}

\section{Conclusion}
In this paper, we present a well-performed frame reconstruction model and a simple key frame selector which make use of the frame reconstructor to reduce the computation costs and increase accuracy. However, our key frame selector only supports a 2 out of 8 selection strategy for extraction, which didn't take the contents of the video into consideration. What's more, diving the video into video clips with 8 frames is not a ideal way to deal with the video content. 

%
%
%

%




\end{document}